# Personalized human mobility prediction for HuMob challenge

Masahiro Suzuki[†]
Graduate Degree Program of
Applied Data Sciences
Sophia University
Chiyoda-ku, Tokyo, Japan
m-suzuki-5y3@eagle.sophia.ac.jp

Shomu Furuta[†]
Graduate Degree Program of
Applied Data Sciences
Sophia University
Chiyoda-ku, Tokyo, Japan
s-furuta-3s7@eagle.sophia.ac.jp

Yusuke Fukazawa[†]
Graduate Degree Program of
Applied Data Sciences
Sophia University
Chiyoda-ku, Tokyo, Japan
fukazawa@sophia.ac.jp

**Abstract**

We explain the methodology used to create the data submitted to HuMob Challenge, a data analysis competition for human mobility prediction. We adopted a personalized model to predict the individual's movement trajectory from their data, instead of predicting from the overall movement, based on the hypothesis that human movement is unique to each person. We devised the features such as the date and time, activity time, days of the week, time of day, and frequency of visits to POI (Point of Interest). As additional features, we incorporated the movement of other individuals with similar behavior patterns through the employment of clustering. The machine learning model we adopted was the Support Vector Regression (SVR). We performed accuracy through offline assessment and carried out feature selection and parameter tuning. Although overall dataset provided consists of 100,000 users trajectory, our method use only 20,000 target users data, and do not need to use other 80,000 data. Despite the personalized model's traditional feature engineering approach, this model yields reasonably good accuracy with lower computational cost.

## 1 Introduction

Predicting people's trajectory in urban areas has become an essential task in fields such as traffic modeling and urban planning. People flow effects several complex traffic tasks such as taxi demand forecast[1] and bike share rebalancing problem[2]. Travel surveys or national census have been used to capture peoples flow in the real world. Recently, mobile phones and social media are used to estimate people's trajectory data. Despite the importance of the problem, it is difficult to benefit from open technological development as the trajectory data is independently retained by each organization from the perspective of privacy.

In such a context, the HuMob challenge[3] is landmark competition as the challenge provides realistic data set of human mobility. Yahoo Japan Co., Ltd., provided the data representing the movement routes of individuals in a major urban area over a period of 90 days. We conducted the prediction of people's movement for two given tasks. For the tasks, we forecasted the movement of 100,000 people under normal circumstances in Task 1, and the movement of 25,000 people in emergency situations in Task 2.

When predicting people's movement patterns, five general approaches can typically be considered.

1) Personalized models using machine learning: each user has own prediction model.

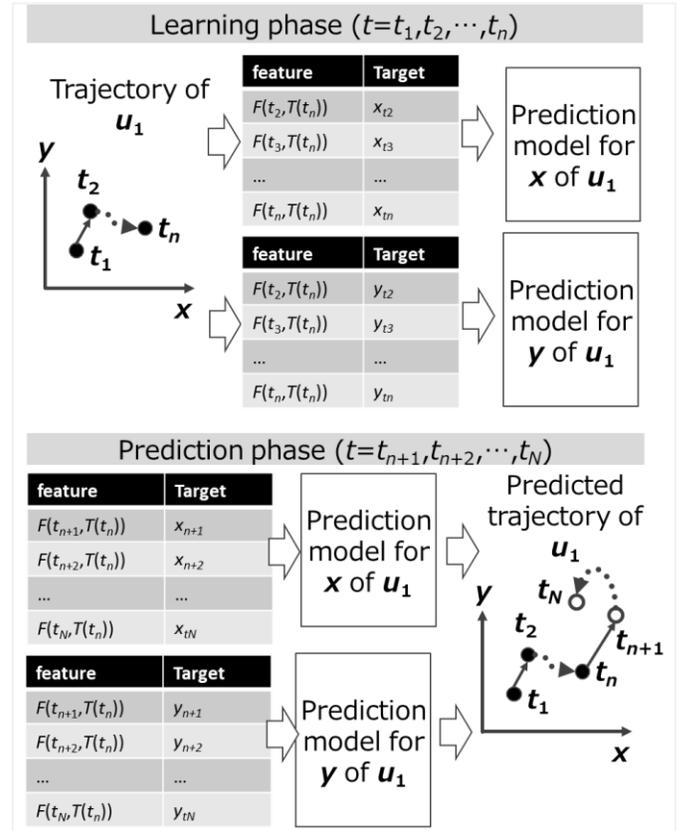

Figure 1 Overview of the proposed method

2) One model using machine Learning: one model explains all users trajectory.
3) Matrix factorization (e.g.: SVD)
4) Deep learning based time-series predictions (e.g.: LSTM)
5) Long sequence time-series forecasting (e.g. Transformer)

Although we would like to try all the above, we adopt approach 1) personalized model considering the time limitation to the deadline. The reason why we choose 1) personalized model is based on the hypothesis that human movements are entirely unique to each individual.

In the following, we describe the proposed method in Section 2. We described offline evaluation and dataset creation for submission in Section 3 and 4, respectively. We conclude this paper in Section 5.

## 2 Proposed Method

Figure 1 illustrates the overview of the proposed method. In this method, we create the personalized model for each user. The model consists of two models for *x* and *y* prediction that is there are 40,000 models for 20,000 users. In the learning phase of user *u*, first we create feature data to predict $x_{t(2)}$. The feature data is generated from the datetime $t_2$ and the trajectory $T(t_n)=\{(x_{t1},y_{t1}), (x_{t(1)},y_{t(1)}),…,(x_{t(n)},y_{t(n)})\}$. We generate the data for $t_2, t_3,…,t_n$. We do the same procedure for *y*. Then we learn the prediction model for *x* and *y* using the dataset. In the prediction phase, we generate feature data from the datetime $t_2$ and the trajectory $T(t_n)=\{(x_{t1},y_{t1}), (x_{t(1)},y_{t(1)}),…,(x_{t(n)},y_{t(n)})\}$. Finally, we predict target trajectory $\{(x_{t(n+1)},y_{t(n+1)}),…,(x_{t(N)},y_{t(N)})\}$ using the learned model to predict future trajectory of user *u*. In the following sections, we describe feature engineering for personalized model.

### 2.1 Feature Design

Table 1 lists the features used in building this model. The processed and generated features were created from the following seven perspectives.

Table 1: Designed features

|   | feature value | # dimensions |
|---|---|---|
| 1 | Date and time | 4 |
| 2 | Activity time | 4 |
| 3 | Days of the week | 7 |
| 4 | Weekday or holiday | 2 |
| 5 | AM/PM | 2 |
| 6 | Frequency to visit POI categories | 85 |
| 7 | clustering | [task1]BC:5,PC:50 [task2]BC:5 |

1) Date and time

We converted date and time to capture the circular nature of date and time. If the date information is used in its original form, we cannot capture the cyclical movement of one week (7 days). Similarly, if time information is used in the range of 0-47, we cannot capture the cyclical movement of 24 hours. In this paper, date and time are represented in two dimensions ($\alpha_d, \beta_d, \alpha_t, \beta_t$) based on the following formula.

$$\alpha_d = \sin\frac{2\pi}{7}d, \beta_d = \cos\frac{2\pi}{7}d, \{0 \leq d < 7\} \quad (1)$$

$$\alpha_t = \sin\frac{\pi}{24}t, \beta_t = \cos\frac{\pi}{24}t, \{0 \leq t < 48\} \quad (2)$$

2) Activity time

We assume that people's behavior patterns change over a time of the day, such as dawn, daytime, evening, and night. For instance, most of the people go out during the daytime, and go to sleep at home in the night. Thus, we discretized the time *t* and generated period of activity. In this paper, we represented the activity time as a four-dimensional ($f_{act}, f_{high\_act}, f_{rest}, f_{deep\_rest}$) based on the following formula.

$$f_{act} = \begin{cases} 1 & \text{if } 18 \leq t < 23, 33 \leq t < 38 \\ 0 & \text{if otherwise} \end{cases} \quad (3)$$

$$f_{high\_act} = \begin{cases} 1 & \text{if } 23 \leq t < 33 \\ 0 & \text{if otherwise} \end{cases} \quad (4)$$

$$f_{rest} = \begin{cases} 1 & \text{if } 13 \leq t < 18, 38 \leq t < 44 \\ 0 & \text{if otherwise} \end{cases} \quad (5)$$

$$f_{deep\_rest} = \begin{cases} 1 & \text{if } 0 \leq t < 13, 44 \leq t < 48 \\ 0 & \text{if otherwise} \end{cases} \quad (6)$$

3) Days of the week

We assume that people's behavior changes depending on the day of the week. We created dummy variables with seven days of the week for date information.

4) Weekday or holiday

We assume that people's behavior patterns change depending on whether it is a weekday or a weekend. For instance, one might go to places that are different from usual, such as shopping centers on weekends. Based on the following formula, the weekday flag ($f_{weekdays}$) was expressed in two dimensions.

$$f_{weekdays} = \begin{cases} 0 & \text{if } day\_of\_week = \{Sat, Sun\} \\ 1 & \text{if otherwise} \end{cases} \quad (7)$$

5) AM/PM

We assume that people change their patterns of behavior before and after lunch. For instance, people stay at home during the morning and go out in the afternoon. We represent the morning and afternoon periods in two dimensions, ($f_{AM}, f_{PM}$) based on the following formula.

$$f_{AM} = \begin{cases} 1 & \text{if } 0 \leq t < 25 \\ 0 & \text{if otherwise} \end{cases} \quad (8)$$

$$f_{PM} = \begin{cases} 1 & \text{if } 25 \leq t < 48 \\ 0 & \text{if otherwise} \end{cases} \quad (9)$$

6) Frequency to visit POI categories

We assume that individual behavior patterns can be characterized from the frequency of the type of stores they visit. For instance, supermarkets are commonly used by many people on a daily basis to purchase groceries. On the other hand, only certain people go to baseball stadiums or concert halls. Therefore, we created features that represents frequency of visit for each POI (Point of Interest) categories. We calculate the feature using the POI category dataset provided by HuMob competition. POI category dataset consists of the number of POIs in each category placed in each location mesh (*x,y*). POI categories have 85 dimensions. We follow following procedures to output the above features.

A) For each user *u* and timestamp *t*, we generate the 85-dimetion vector $f_{POI}[c]$=poicount(*u,t*,c) that represents the number of POIs for each POI category c at the mesh ($x_{u,t},y_{u,t}$). We fill zero to the POI category that does not exist. The function poicount($a_1,a_2,…,a_n$) returns the number of POIs for each elements $a_1,a_2,…,a_n$.

B) For each user *u*, we update the vector by $f_{POI}[n]$= poicount(*u,fv*,c). We aggregate the occurrence count of POI

Personalized human mobility prediction for HuMob challenge categories by four activity time features $fv=\{f_{act}, f_{high\_act}, f_{rest}, f_{deep\_rest}\}$.

C) For each user $u$, we update the vector by $f_{POI}[n]=$ poicount($u,fv,c$)/poicount($u$). poicount($u$) represents the total occurrence count of all POI categories for each user $u$.

7) Clustering

We assume that individuals can be clustered into several types of similar behavioral patterns. For instance, two individuals exhibit similar behavior by working at the closely located company and adhering to the same working hours. We conducted both hard and soft clustering against several feature sets. We prepared two feature sets B: Basic Features (Features 2-5) and POI Features (Feature 6). We use *k*-means and the Gaussian Mixture Model (GMM) as a clustering methods. In the case of *k*-means, we calculated the cluster number for each user and then transformed into feature quantities by dummy variable transformation. In the GMM, we assign the occurrence probability of each cluster to each user.

## 2.2 Learning Model

In this study, we tested several machine learning models and choose Support Vector Regression (SVR) with rbf kernel. The SVR was found to provide better accuracy when compared to these models such as Random Forest, LightGBM and XGBoost. Considering the computational time to select features and a substantial volume of the data used in this study, we have to select the models that offered a small computational cost.

Table 2: Feature Selection for Task 1

| | Features | | | | Accuracy Evaluation | | | |
|---|---|---|---|---|---|---|---|---|
| | B | P | BC | PC | GEO | SDG | DTW | SDD |
| 1 | ✓ | | | | 0.23 | 0.18 | 50.13 | 51.88 |
| 2 | | ✓ | | | 0.20 | 0.18 | 63.08 | 69.39 |
| 3 | ✓ | ✓ | | | 0.23 | 0.18 | 50.30 | 52.27 |
| 4 | | | ✓ | | 0.22 | 0.18 | 57.33 | 62.89 |
| 5 | | | | ✓ | 0.16 | 0.18 | 61.96 | 71.01 |
| 6 | ✓ | | ✓ | | 0.23 | 0.18 | 50.20 | 52.51 |
| 7 | | ✓ | | ✓ | 0.17 | 0.18 | 62.29 | 71.03 |
| 8 | | ✓ | ✓ | | 0.22 | 0.18 | 57.32 | 62.82 |
| 9 | ✓ | | ✓ | ✓ | 0.23 | 0.18 | 50.46 | 52.90 |
| 10 | ✓ | ✓ | ✓ | | 0.23 | 0.18 | 50.43 | 52.84 |
| 11 | ✓ | ✓ | ✓ | ✓ | 0.23 | 0.18 | 50.82 | 53.25 |

Table 3: Feature Selection for Task 2

| | Features | | | | Accuracy Evaluation | | | |
|---|---|---|---|---|---|---|---|---|
| | B | P | BC | PC | GEO | SDG | DTW | SDD |
| 1 | ✓ | | | | 0.22 | 0.22 | 55.30 | 58.28 |
| 2 | | ✓ | | | 0.24 | 0.23 | 51.98 | 65.16 |
| 3 | ✓ | ✓ | | | 0.22 | 0.22 | 55.13 | 58.16 |
| 4 | | | ✓ | | 0.24 | 0.22 | 53.22 | 58.24 |
| 5 | | | | ✓ | 0.19 | 0.22 | 49.88 | 58.10 |
| 6 | ✓ | | ✓ | | 0.22 | 0.22 | 55.41 | 59.00 |
| 7 | | ✓ | | ✓ | 0.20 | 0.22 | 50.34 | 58.74 |
| 8 | | ✓ | ✓ | | 0.24 | 0.22 | 53.25 | 58.37 |
| 9 | ✓ | | ✓ | ✓ | 0.22 | 0.22 | 55.28 | 58.90 |
| 10 | ✓ | ✓ | ✓ | | 0.22 | 0.22 | 55.25 | 58.84 |
| 11 | ✓ | ✓ | ✓ | ✓ | 0.22 | 0.22 | 55.21 | 58.75 |

## 3 Offline Evaluation

We perform the offline evaluation for feature selection and parameter tuning. From a dataset of 100,000 individuals, we extracted data for 20 people and 100 people who were not masked. We mask the extracted dataset from day 60 to day 74. We tune the model based on four evaluation metrics: average GEOBLEU (GEO), standard deviation of GEOBLEU (SDG), average DTW (DTW), and standard deviation of DTW (SDD). The detail definition of GEOBLEU and DTW are shown in the paper[4].

We compare the following four feature sets using 20 people datasets.
- B: Basic Features (Features 2-5 in Selection 2.1)
- P: POI Features (Features 6 in Selection 2.1)
- BC: Clustering features of the basic features
- PC: Clustering features of the POI features

The clustering numbers for BC and PC were set to five. We created models using various combinations of B, P, BC, and PC, and conducted accuracy evaluations. The evaluation results for Task 1 are shown in Table 2. According to these results, the combination in pattern 9 (B, PC, PC) had the best accuracy, hence, we adopted pattern 9. The evaluation results for Task 2 are presented in Table 3. According to these results, Pattern 2 had the highest accuracy and was therefore adopted.

We then tuned the cluster number using 100 people datasets. In pattern 9, we set the BC cluster number to 5 and 10, and the PC cluster number to 10, 50, 100, and 150, and performed accuracy evaluations. The results showed that for Task 1, the highest accuracy was achieved when the BC cluster number was 5 and the PC cluster number was 50. For Task 2, the highest accuracy was achieved when the BC cluster number was set to 5. We compared the *k*-means and GMM methods for clustering. The results showed that *k*-means had higher accuracy for both tasks, and thus, we adopted the *k*-means method.

## 4 Data for Submission

Based on the offline evaluation, for Task 1, we constructed submission models using the fundamental features of Features 1-5, clustering features of the basic features (5 clusters), and clustering features related to POI (50 clusters). For Task 2, we constructed submission models using clustering results of the basic features (5 clusters). Although overall dataset provided consists of 100,000 users trajectory, our method use only 20,000 target users data, and do not need to use other 80,000 data. The accuracy evaluation results notified from the organizers are as follows.

|  | Task 1 (GEO) | Task 1 (DTW) | Task 2 (GEO) | Task 2 (DTW) |
|---|---|---|---|---|
| Score | 0.23416211 | 31.73553402 | 0.14653886 | 45.03996966 |

## 5 Conclusion

In this paper, we utilized personalized models for human mobility prediction tasks. Despite the personalized model's traditional feature engineering approach, this model yields reasonably good accuracy with lower computational cost. In the feature engineering process, we found that the clustering feature and POI-frequency feature are important. As a future work, we would like to consider the following methods such as deep learning based time-series predictions (e.g.: LSTM) and Long sequence time-series forecasting (e.g. Transformer).

### Acknowledgements

This work was supported by Sophia University Special Grant for Academic Research. We would like to thank Prof. Tomi Ohtsuki, Sophia University for providing the computing environment for our participation in this competition.